\documentclass[10pt,twocolumn,letterpaper]{article}
\usepackage[pagenumbers]{cvpr}
\usepackage{booktabs}
\newcommand{\SEL}{\mathrm{SEL}}
\newcommand{\MON}{\mathrm{MON}}
\newcommand{\SELadd}{\mathrm{SEL}_{\mathrm{add}}}
\newcommand{\armOracle}{Oracle}
\newcommand{\armSeg}{Det{+}SAM}
\newcommand{\armRaw}{SAM-raw}
\newcommand{\armPatch}{Patch}
\newcommand{\ea}{\textsc{ea}}
\newcommand{\vo}{\textsc{vo}}
\definecolor{cvprblue}{rgb}{0.21,0.49,0.74}
\usepackage[pagebackref,breaklinks,colorlinks,allcolors=cvprblue]{hyperref}
\hypersetup{
  pdftitle={When Do Frozen VLMs Respond to Image-Free Object-Token Edits?},
  pdfauthor={Wonbin Son; Gyumun Choi; Junil Seo; Seungmin Rho; Mi Young Lee; Hyungjoon Kim}
}

\title{When Do Frozen VLMs Respond to Image-Free Object-Token Edits?}

\author{%
Wonbin Son$^{1,*}$\hspace{0.6em}
Gyumun Choi$^{1}$\hspace{0.6em}
Junil Seo$^{1}$\hspace{0.6em}
Seungmin Rho$^{2}$\hspace{0.6em}
Mi Young Lee$^{2}$\hspace{0.6em}
Hyungjoon Kim$^{1,\dagger}$\\
{\normalsize $^{1}$Changwon National University\quad
$^{2}$Chung-Ang University\quad
$^{*}$First author\quad $^{\dagger}$Corresponding author}\\
{\footnotesize\ttfamily
\{diwjidghk78, cgyumun, junil0106\}@gmail.com\quad
\{smrho, miylee\}@cau.ac.kr\quad hyungjoon@changwon.ac.kr}%
}

\begin{document}
\maketitle
% Abstract: approved English v4 with the confirmed title, 2026-10-26.
% Frozen visual encoder and language backbone; only the projector is trained.
\begin{abstract}
Answering what-if queries about a scene with a vision-language model (VLM) usually involves stating the assumption in text or editing the scene with a generative model. We instead move the edit to the representation level, before the model input. The image is represented as a set of object tokens, and the model receives edited tokens without accompanying patch tokens from the original image. This design rests on an open question---when do frozen VLMs actually respond to image-free object-token edits? We study this question in a simple testbed, keeping the visual encoder and language backbone frozen and training only a projector. Our analysis reveals three findings. \textbf{Responsiveness depends on training:} unlike standard VQA training, explicit edit training produces responsiveness in dense scenes and strengthens it in sparse scenes, across deletion, movement, and addition. \textbf{Token construction and context token count shape responsiveness:} the observed differences between token constructions appear to reflect more than token count alone, and responsiveness tends to weaken as surrounding context grows. \textbf{Standard VQA performance and edit responsiveness are distinct:} the object-token route preserves $92$--$96\%$ of a matched patch-token baseline's free-form VQA accuracy, and responses depend on the input tokens. The observed trends in responsiveness, token construction, and VQA retain their direction across two remote-sensing datasets (iSAID, VRSBench) and three frozen language backbones. We release the evaluation-question generator, measurement records, scoring logs, and code.
\end{abstract}

% Research question aligned with the confirmed title, 2026-10-26.
% Applied 2026-10-24: approved Introduction English draft v6.
% Figure 1 caption follows the previously agreed Korean v2 caption.
% Original WACV submission remains preserved in outputs/paper/.
\section{Introduction}
\label{sec:intro}

\begin{figure*}[t]
  \centering
  \includegraphics[width=\textwidth]{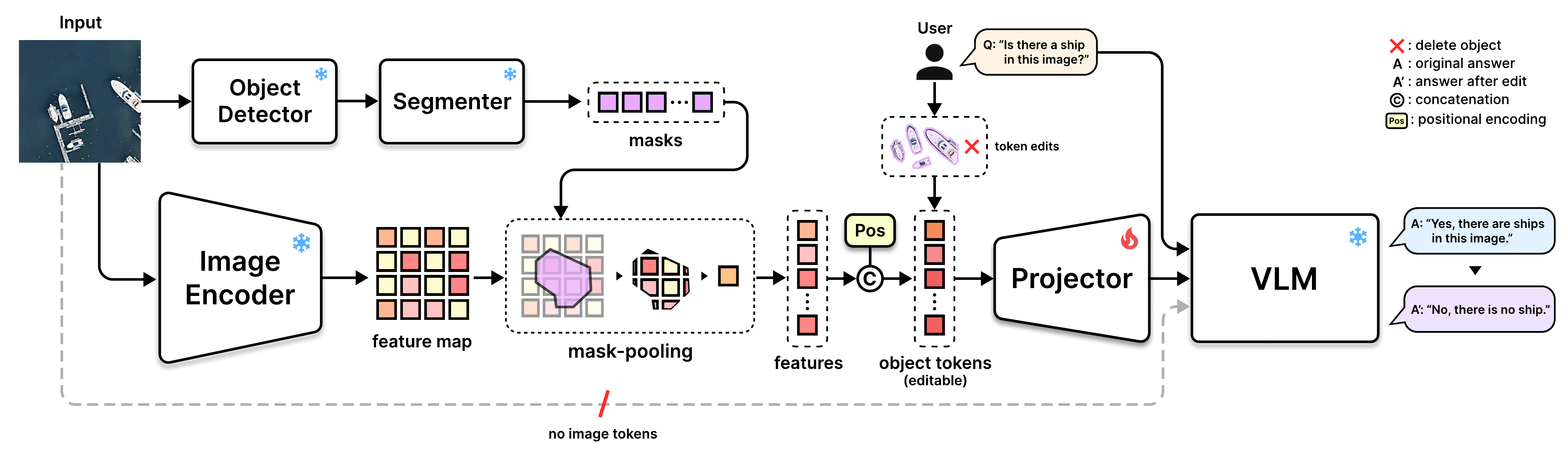}
  \vspace{-6mm}
  \caption{\textbf{Object-token construction and editing.} Features are pooled over object regions from a frozen visual encoder's feature map and combined with positional information to form object tokens. A projector maps these tokens into the input space of a frozen language backbone. Object tokens are edited before model input, without accompanying patch tokens from the original image. Only the projector is trained.}
  \label{fig:framework}
\end{figure*}

Hypothetical reasoning about a scene considers changes to the observed scene and asks about their consequences. For example, in satellite and aerial imagery, one might ask whether any ships would remain in a harbor if selected ships were removed. Such questions can support planning and scenario analysis, and systems for answering hypothetical questions have been proposed in remote sensing~\cite{elgendy2025chatenv}. Existing approaches convey the hypothetical changes through text~\cite{zhang2023cvqa} or answer questions using images edited by generative models~\cite{agarwal2020ivvqa}.

We take a different perspective. A model's answer alone does not directly reveal how a hypothetical change has been incorporated into its reasoning. We therefore move the edit to the representation level, before the model input, to examine model responses to explicitly specified input changes. We represent an image as a set of tokens encoding object features and positions, explicitly apply the hypothetical edit to this set, and then feed it to a vision-language model (VLM). This approach rests on an important premise: a model with frozen backbones must be able to adjust its responses appropriately when object tokens are edited. For example, if the token representing the only ship in a region is deleted but the model still answers that a ship is present there, the edited state is not correctly reflected in the final answer. We ask under what conditions this premise holds: \textbf{when do frozen VLMs respond to image-free object-token edits?}

We study how edit responsiveness varies with training, as well as the characteristics of the observed responses and the conditions under which they arise. To this end, we construct a simple testbed that directly edits object tokens and measures the direction and magnitude of changes in answer probabilities (\cref{fig:framework}). We pool features over object regions from a frozen visual encoder's feature map, combine them with positional information, and feed them through a projector to a frozen language backbone. In the main experiments, only a two-layer projector with approximately 15 million parameters is trained. Deletion removes tokens, movement modifies their positional information, and addition inserts features extracted from real objects. We conduct analyses on two remote-sensing datasets and repeat selected comparisons of training and token construction across three language backbones. The observations from this testbed form the basis of three contributions:
\begin{enumerate}
  \item \textbf{We show how training conditions shape edit responsiveness.} Learning to answer questions from object tokens through standard visual question answering (VQA) training does not guarantee responsiveness to edits. Training that includes edited examples strengthens responsiveness, and we analyze how its effects vary across deletion, movement, and addition.
  \item \textbf{We systematically characterize edit responsiveness.} The analysis covers object-token construction and token count, selected comparisons across language backbones and model sizes, and the relationship between edit responsiveness and standard VQA performance.
  \item \textbf{We provide resources for measurement and analysis.} We release code, evaluation-question generation rules, and measurement records to support reproducible analyses of edit responsiveness and follow-up research.
\end{enumerate}

This characterization may serve as a design guide for object-token editing systems and related research, informing choices about edit training, token construction, and the amount of surrounding context.

% CVPR Related Work, approved 2026-10-24; source: drafts/2026-10-24-related-work-v7-en.tex.
\section{Related Work}
\label{sec:related}

\paragraph{Counterfactual queries and answer reliability.}
Visual question answering about hypothetical scene changes has been studied through textual premises and edited image inputs~\cite{zhang2023cvqa,agarwal2020ivvqa}. MindEdit-Bench~\cite{yu2026mindedit} evaluates reasoning about spatial relations and visibility after hypothetical object movements or rotations. With textual premises, the input image still depicts the observed scene, leaving the model to account for the hypothetical change when answering. A related line examines dependence on visual evidence through language-bias separation~\cite{niu2021cfvqa}, substitution of question-irrelevant object features~\cite{gupta2022swapmix}, and hallucination diagnostics before and after object removal~\cite{yebin2024beaf}. These studies motivate examining both hypothetical reasoning and whether responses reflect input changes. We apply edits directly to object tokens before model input, measuring the direction and magnitude of changes in answer probabilities to characterize responsiveness and the conditions under which it arises.

\paragraph{Object-token inputs to language models.}
Object-level language-model inputs have been explored across domains, including autonomous driving and 3D scene understanding~\cite{tian2024tokenav,huang2024chatscene}. For 2D images, related approaches combine mask-based object representations with global and patch tokens, compress tokens at the object level, or connect mask-specified regions with language instructions~\cite{jahagirdar2026maskllava,zhang2025adatok,yuan2024osprey}. Building on this foundation, we treat object tokens as editable input units and examine model responses to their modification. In addition, work on semantic tokenization addresses how excessive fragmentation can impair visual semantic integrity~\cite{wu2024setok}, informing our analysis of the relationship between token construction and edit responsiveness.

\paragraph{Intervention-aware learning.}
In concept bottleneck models, allowing interventions does not guarantee appropriate predictive responses. IntCEM~\cite{zarlenga2023intcem} notes that standard objectives may offer insufficient incentives to use interventions and incorporates intervention response into training. Related studies formalize and optimize intervention effects~\cite{laguna2024intervenable}, mitigate hallucinations through internal representation edits~\cite{jiang2025interpreting}, and assess explanation faithfulness through image edits~\cite{ding2025edct}. We apply this training perspective to object-token inputs. With frozen backbones, we compare VQA training with and without edited examples to quantify responsiveness across training and input conditions.

\paragraph{Remote-sensing vision-language models.}
Remote-sensing VLMs address region-referenced question answering, fine-grained relationship understanding, and stepwise geospatial reasoning~\cite{kuckreja2024geochat,luo2024skysensegpt,liu2025rsthinker}. ChatENV~\cite{elgendy2025chatenv} supports hypothetical question answering in remote-sensing scenarios. Prompt-RSVQA~\cite{chappuis2022promptrsvqa} converts visual context into text for language-model input, providing a precedent for alternative representations of imagery. Such hypothetical questions also arise in remote sensing, where we study how models respond to object-token edits.

% Section 3 applied draft v7, 2026-10-24.
% Combines the movement comparison and MON definition in the approved sentence.
\section{Testbed and Measurements}
\label{sec:method}

\subsection{Object-token construction and editing}
\label{sec:interface}

We represent an image as a set of individually editable object tokens. A frozen visual encoder first produces a spatial feature map. For each object mask, we average the features within its region to obtain an object feature~\cite{li2025maskadapter}. To reduce the component shared across the image, we subtract a fraction $\lambda$ of the mean feature over the image. We use $\lambda=0.5$ by default (with masks obtained by detection followed by segmentation, deletion responsiveness was slightly lower at $\lambda=0$ and similar across the other tested values; supplementary). Each token also receives positional information encoding its centroid, bounding box, and area. No class label is attached to the token; the detector's class predictions are used only to propose masks. Object tokens enter the language model as a single sequence, without accompanying patch tokens from the original image. Implementation details, including positional encoding, are provided in the supplementary material.

We vary the mask source as an experimental condition, using ground-truth instance masks, masks obtained by detection followed by segmentation, and the full set of masks returned directly by a segmenter. These choices jointly affect mask precision, token count, the extent to which an object is split across tokens, and which objects are represented. We therefore examine token characteristics alongside edit responsiveness. We apply three edits to the resulting token set before it enters the language model. \textbf{Deletion} removes the target object's token. \textbf{Movement} changes its positional information while preserving its features. \textbf{Addition} copies features extracted from a real object in the same or another scene and inserts them with new positional information. Using object tokens as the editing unit, we measure how these changes are reflected in model responses.

\subsection{Teaching edits to a frozen VLM}
\label{sec:teaching-method}

To compare edit responsiveness under different training conditions, we freeze the visual encoder and language model and train only a projector that maps object features into the language model's input space. The projector is a two-layer multilayer perceptron (MLP) with approximately 15 million trainable parameters in the main configuration. Using the same architecture, we compare standard visual question answering (VQA) training alone with VQA training that also includes deletion and movement examples. We also use a condition that includes addition training and distinguish these settings in the experiments. This simplicity is a deliberate choice for comparing edit responsiveness across training and input conditions. Our contribution lies in the systematic characterization of the responses observed in this testbed.

We automatically construct training examples from edit rules, pairing token sets before and after an edit with their corresponding target responses. For example, deletion training asks whether an object is present in a specified region and targets \emph{no} after the corresponding object token is removed. Movement training relocates an object across the image midline and targets a scene description reflecting its updated location. Both vertical and horizontal movements are included. Addition training uses three cases: \emph{no} for a region without the target class, \emph{yes} after inserting real-object features of that class, and \emph{no} maintained after inserting features of a different class.

We use a fixed class split and generate edit-training examples for only one half. Edits on both halves are evaluated regardless of whether their classes appear in VQA training. Edit examples are mixed with VQA examples; the proportions of edit types are selected in an initial sweep and fixed across comparisons. Exact ratios and training settings are in the supplementary.

\subsection{Measuring edit responsiveness}
\label{sec:protocol}

We construct evaluation questions and expected response changes using existing object class and position annotations together with the edit rules. For deletion and addition, the edit determines whether the queried class is present in the specified region, so no separate manual annotation of post-edit answers is required. We also apply a \textbf{placebo edit}: the same type of operation performed in a way that should leave the answer unchanged. We subtract its effect to distinguish responses to question-relevant edits from general changes caused by modifying the input.

Each question specifies an answer pair: \emph{yes/no}, \emph{top/bottom}, or \emph{left/right}. We compute the relative probabilities of the two candidates from the logits at the first decoding step:
\[
P(a\mid Z,q)=
\frac{\exp \ell_a}{\sum_{a'\in\mathcal A_q}\exp \ell_{a'}},
\qquad
\ell_a=\max_{t\in\mathcal T_a}z_t.
\]
Here, $Z$ is the object-token set, $q$ is the question, and $\mathcal A_q$ is the pair of candidate answers. The logit $z_t$ is the model's score for the next output token $t$. The set $\mathcal T_a$ contains the first-token IDs of capitalization and leading-space variants of answer $a$ under the model's tokenizer; we use the highest score among them. No sampling or temperature scaling is applied. We evaluate cases in which the correct answer before the edit has probability greater than $0.5$, and measure changes in this relative probability after editing. Each metric below is computed per case and then averaged over the evaluated cases.

\textbf{Deletion (SEL).} Removing the target object from the queried region determines the presence answer to be \emph{no}. We compute $\SEL=\Delta P_{\mathrm{rel}}-\Delta P_{\mathrm{irrel}}$, where $\Delta P_{\mathrm{rel}}$ and $\Delta P_{\mathrm{irrel}}$ are the decreases in $P(\mathrm{yes})$ after deleting target tokens and question-irrelevant tokens, respectively. \textbf{Movement (MON).} We compare moves across the image midline with moves within the same half using $\MON=\Delta P_{\mathrm{cross}}-\lvert\Delta P_{\mathrm{null}}\rvert$, computed separately for the vertical and horizontal axes. Here, $\Delta P_{\mathrm{cross}}$ is the change in the probability assigned to the destination side after crossing the midline, and $\Delta P_{\mathrm{null}}$ is the change in that same answer's probability under the placebo move. If the scene contains only one object of the queried class, crossing the midline changes the correct answer to the position question. Evaluation also includes cases with multiple objects of that class, where MON measures response sensitivity to the target object's crossing of the midline. \textbf{Addition ($\SELadd$).} Inserting features of the target class into a region without that class determines the presence answer to be \emph{yes}. Let $\Delta P$ be the resulting increase in $P(\mathrm{yes})$ and $\Delta P_{\mathrm{placebo}}$ the increase after inserting features of a different class. We compute $\SELadd=\Delta P-\Delta P_{\mathrm{placebo}}$. A control that changes the insertion location is examined in the supplementary material.

Separately, for free-form VQA, we report $\mathrm{driven}=\mathrm{acc}_{\mathrm{normal}}-\mathrm{acc}_{\mathrm{shuffled}}$, using a fixed permutation of entire object-token sets across evaluation images while keeping questions unchanged. The same token sets are reused, so the accuracy difference reflects dependence on their correct pairing with questions. We interpret this control alongside the edit metrics in the experiments.

% Section 4 applied draft v6, 2026-10-26.
% Approved prose; scoring audit moved last; original numeric table entries retained.
\section{Experiments}
\label{sec:experiments}

\subsection{Setup}
\label{sec:setup}

We evaluate edit responsiveness on two remote-sensing datasets. On \textbf{iSAID} \cite{zamir2019isaid,xia2018dota}, we use 374 full images from the validation set and standard VQA questions from RSVLM-QA \cite{zi2025rsvlmqa} over the same imagery. The median token count per image is approximately 47 with ground-truth masks and 1,005 with masks returned directly by SAM, supporting analysis of scenes with many objects and tokens. The main evaluation uses full images with up to 4,096 tokens each. A separate analysis of token construction uses 459 dense 512$\times$512 tiles containing at least ten objects. The tile experiment varying token count uses 729 tiles from 33 source images, with its conditions described in \cref{sec:cleanliness-density}. \textbf{VRSBench} \cite{li2024vrsbench} provides 1,121 tiles of 512$\times$512 pixels and short-answer VQA. Its average of approximately 1.8 objects per tile provides a sparse setting. Edit-evaluation questions for both datasets are generated programmatically from object annotations and edit rules. Our analysis requires both object-level annotations, for constructing ground-truth-based tokens and examining token construction and count, and VQA, for assessing preservation of standard question-answering performance. These requirements limit the available datasets: VRSBench provides both, while iSAID meets them when paired with RSVLM-QA over the same imagery.

We compare three token constructions: \textbf{Oracle}, \textbf{Det+SAM}, and \textbf{SAMraw}. Oracle uses ground-truth object annotations: instance masks on iSAID, and masks obtained by prompting SAM \cite{kirillov2023sam} with ground-truth oriented boxes padded by 15\% on VRSBench. Det+SAM prompts SAM with boxes from YOLO-OBB detections \cite{jocher2024yolo11}, whereas SAMraw uses the full set of masks returned directly by SAM. The \textbf{Patch} reference for standard VQA pools the same feature map on a 16$\times$16 grid to produce 256 tokens. We match the encoder, projector, and VQA training conditions to compare the regions over which features are pooled. Our base training conditions are \textbf{VO}, using standard VQA alone, and \textbf{EA}, combining VQA with deletion and movement examples. Training that also includes addition is distinguished in \cref{sec:teaching}.

The default language backbone is Qwen2-7B from LLaVA-OneVision-7B \cite{li2024llavaonevision}. We extract object features from a frozen ConvNeXt \cite{ilharco2021openclip,liu2022convnext} at an input resolution of 512 pixels and train only a two-layer MLP projector with approximately 15M parameters. Encoder comparisons and the rationale for this choice are provided in the supplementary. SigLIP \cite{zhai2023siglip}, LLaVA-OneVision's native visual encoder, is used for image-input comparisons, including image copresence (\cref{sec:reading}), not for object tokens. Comparisons with other language backbones and a 0.5B model are presented in \cref{sec:universality}.

\paragraph{Statistics.} Unless stated otherwise, main results aggregate nine runs across three folds and three seeds, with 95\% confidence intervals from 5,000 percentile-bootstrap resamples of per-run means. For condition contrasts, we resample these means independently unless marked as paired. Full-image training and evaluation use disjoint source images; \cref{sec:cleanliness-density} details the tile-density split and image-level resampling for density confidence intervals.

\paragraph{Training budget.} We fix training settings, including epochs and learning rate, without per-condition tuning or convergence-based early stopping. These settings support controlled comparisons but do not imply optimal performance. We focus on relative differences, confidence intervals, and consistent directions across repeated comparisons rather than absolute performance or responsiveness magnitudes.

\paragraph{Scoring and reproducibility resources.} Edit responsiveness uses pre-edit correctness and relative probabilities of two answer candidates (\cref{sec:protocol}). We extract counts from free-form responses and score standard VQA with a fixed Qwen2.5-72B judge \cite{qwen2024qwen25}. Two independent annotators agreed on 143 of 150 stratified responses. On these 143 responses, also used to refine the rubric, judge agreement was 91.6\% (Cohen's $\kappa=.83$). Inter-annotator agreement across all 150 responses was 95.3\% ($\kappa=.90$). We release question-generation code based on object annotations and edit rules, prompts, per-run scores and evaluation counts, exclusion rules, scoring rubrics, and logs. Implementation, hardware, training settings, statistical procedures, per-run results, and additional analyses are in the supplementary.

\paragraph{Why remote sensing.} Remote sensing combines a demand for questions about hypothetical scene changes with the conditions needed for our analysis (\cref{sec:intro}). Instance annotations in dense scenes support tokens based on object ground truth and analyses of responsiveness across token constructions and counts. Overhead views also reduce mutual occlusion between objects, facilitating these analyses.

\subsection{VQA training alone does not guarantee edit responsiveness}
\label{sec:notdefault}

\begin{table*}[t]
  \centering\small
  \caption{\textbf{Edit responsiveness by token construction and training ($\SEL$/$\MON$).} $\MON$ is shown as the mean; $\star$ marks a 95\% CI above zero.}
  \label{tab:main}
  \footnotesize\setlength{\tabcolsep}{3.5pt}
  \begin{tabular}{llcccccc}
    \toprule
    & & \multicolumn{3}{c}{iSAID (374 full images)} & \multicolumn{3}{c}{VRSBench (1{,}121 tiles)} \\
    \cmidrule(lr){3-5}\cmidrule(lr){6-8}
    Arm & Train & $\SEL$ & $\MON_V$ & $\MON_H$ & $\SEL$ & $\MON_V$ & $\MON_H$ \\
    \midrule
    \armOracle & \ea & $\mathbf{+.049}$ {\scriptsize$[.043,.057]$} & $+.016^\star$ & $+.011^\star$ & $\mathbf{+.454}$ {\scriptsize$[.436,.471]$} & $+.658^\star$ & $+.625^\star$ \\
    \armOracle & \vo & $+.002$ {\scriptsize$[-.000,.004]$} & $-.027$ & $-.021$ & $+.148$ {\scriptsize$[.140,.154]$} & $+.196^\star$ & $+.196^\star$ \\
    \armSeg & \ea & $\mathbf{+.044}$ {\scriptsize$[.034,.054]$} & $+.034^\star$ & $+.004$ & $\mathbf{+.510}$ {\scriptsize$[.496,.526]$} & $+.521^\star$ & $+.491^\star$ \\
    \armSeg & \vo & $+.003$ {\scriptsize$[.000,.005]$} & $-.024$ & $-.020$ & $+.085$ {\scriptsize$[.080,.090]$} & $+.102^\star$ & $+.076^\star$ \\
    \armRaw & \ea & $-.000$ {\scriptsize$[-.001,.000]$} & $-.011$ & $-.011$ & $+.064$ {\scriptsize$[.051,.076]$} & $+.113^\star$ & $+.051^\star$ \\
    \armRaw & \vo & $+.000$ {\scriptsize$[-.001,.001]$} & $-.013$ & $-.013$ & $-.000$ {\scriptsize$[-.001,.001]$} & $-.012$ & $-.025$ \\
    \bottomrule
  \end{tabular}
\end{table*}

\textbf{In dense scenes, learning to answer questions from object tokens does not by itself yield substantial responsiveness to edits of those tokens.} Under VO, deletion $\SEL$ on iSAID is near zero: $.002$ for Oracle and $.003$ for Det+SAM (\cref{tab:main}). This is not simply a failure to read object tokens. These adapters retain input-token dependence ($\mathrm{driven}>0$, \cref{sec:reading}). Good reading does not guarantee edit responsiveness. We next examine how edit training changes this response.

\subsection{Edit training strengthens responses to all three edits}
\label{sec:teaching}

\begin{table}[t]
  \centering\footnotesize
  \caption{\textbf{Addition responsiveness across token constructions.} L2 copies a same-scene instance's features; L3 uses features of the same class from another scene. The no-add condition is trained on deletion and movement.}
  \label{tab:add}
  \setlength{\tabcolsep}{2pt}
  \scriptsize
  \begin{tabular}{@{}lcccc@{}}
    \toprule
    & \multicolumn{2}{c}{$\SELadd$ L2} & L3 & Untaught \\
    \cmidrule(lr){2-3}
    Arm (9 cells) & taught & no add & taught & classes \\
    \midrule
    \armOracle & $+.195$ {\scriptsize$[.179,.210]$} & $+.050$ & $+.169$ & $+.038$ {\scriptsize$[.013,.064]$} \\
    \armSeg & $\mathbf{+.122}$ {\scriptsize$[.093,.153]$} & $+.033$ & $\mathbf{+.099}$ & $+.022$ {\scriptsize$[-.003,.044]$} \\
    \armRaw & $+.076$ {\scriptsize$[.057,.093]$} & $-.002$ & $+.066$ & $-.004$ {\scriptsize$[-.015,.008]$} \\
    \bottomrule
  \end{tabular}
\end{table}

\textbf{Training with edit examples strengthens responsiveness to all three operations.} Under the same token constructions and training budget, EA increases deletion $\SEL$ over VO by $.047$ for Oracle and $.041$ for Det+SAM on iSAID, with confidence intervals above zero (\cref{tab:main}). Under EA, vertical and horizontal $\MON$ are also mostly positive for Oracle and Det+SAM; the interval for horizontal movement with Det+SAM in dense scenes includes zero. On VRSBench, VO already yields deletion $\SEL$ of $.148$ for Oracle and $.085$ for Det+SAM; EA raises the latter to $.510$ (a $.425$ increase), strengthening an existing response.

We distinguish three addition cases by the source of the inserted features. L1 reinserts a deleted object at its original location as a basic sanity check, with results in the supplementary. \cref{tab:add} covers copying features of an object from the same scene into a region without the target class (L2) and importing features of the same class from another scene (L3). For Det+SAM, addition training raises L2 $\SELadd$ from $.033$ to $.122$ and yields $.099$ in L3. Including addition training increases deletion $\SEL$ and movement $\MON$ by $.073$ and $.032$, respectively. Applying the same training recipe on VRSBench yields an addition $\SEL$ of $.244$ [$.212$, $.274$]. For classes excluded from edit training, addition responses are generally small and their intervals include zero, except for Oracle, where a positive response of $.038$ is observed. These results suggest that addition responsiveness is more restricted to the classes used in edit training than deletion or movement responsiveness.

\subsection{Token construction and context token count shape edit responsiveness}
\label{sec:cleanliness-density}

\begin{figure}[t]
  \centering
  \includegraphics[width=\linewidth]{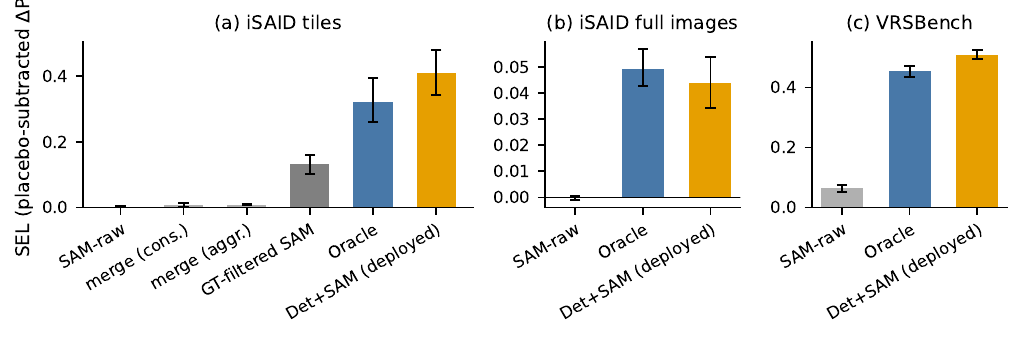}
  \caption{\textbf{Token construction and deletion responsiveness ($\SEL$).} (a) iSAID dense tiles, (b) iSAID full images, (c) VRSBench. Bars: means; error bars: bootstrap 95\% CIs. Each panel uses its own $y$-scale.}
  \label{fig:cleanliness}
\end{figure}

\begin{figure}[t]
  \centering
  \includegraphics[width=\linewidth]{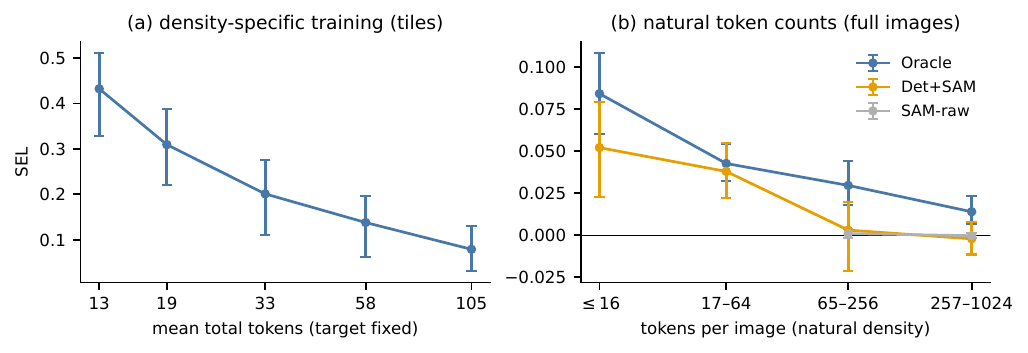}
  \caption{\textbf{Responsiveness and token count.} (a) Tile experiment with a fixed edit target and a separately trained projector at each context-count setting. (b) Full images grouped by natural token count. Error bars in both panels are source-image bootstrap 95\% CIs.}
  \label{fig:density}
\end{figure}

\textbf{After edit training, responsiveness varies with how object tokens are constructed.} On dense iSAID tiles, deletion $\SEL$ increases from $.002$ for SAMraw to $.408$ for Det+SAM (\cref{fig:cleanliness}). Intermediate constructions that merge SAM masks by overlap and feature similarity do not reach the responsiveness of the detection-based construction. Conservative merging does not sufficiently reduce token count, while aggressive merging combines different objects into a single token, making them difficult to edit individually. These results support our choice to construct tokens from detections so that individual objects can be edited. The tendency for ground-truth- or detection-based constructions to outperform SAMraw also appears on full images and VRSBench, and for addition (\cref{tab:main,tab:add}). On full iSAID images, the Oracle--Det+SAM difference is $.005$ [$-.007$, $.018$], with no clear difference detected; on VRSBench, Det+SAM has higher deletion $\SEL$.

Token count must also be considered when interpreting these comparisons. Although Det+SAM scores higher than Oracle at the tile level, Oracle is at least as high when token counts are matched. Token count does not, however, explain all differences between constructions. Retaining only SAM masks matched to ground-truth objects gives approximately 33 tokens per tile, close to Oracle's 36, but $\SEL$ is $.131$ [$.103$, $.159$], compared with $.321$ [$.260$, $.394$] for Oracle.

Mean deletion responsiveness decreases with increasing context token count on iSAID (\cref{fig:density}a). With the edit target fixed, we train a separate projector at each context count, so the curve compares density-specific training conditions. Training and evaluation use different tiles but share source images, which are resampled for confidence intervals. Full-image groups with higher natural token counts likewise show lower responsiveness (\cref{fig:density}b), reflecting differences between images. Attenuation also varies across encoders, whose rankings by object-feature distinctiveness and standard recognition performance differ.

We also examine this trend with the projector held fixed. For each of 73 source images using Oracle tokens, we assign the model from a fold that excluded that image from training. Within each image, we keep the model, target, and question fixed while varying token count over 8, 16, 32, and 64. Among the 23 source images satisfying the pre-edit correctness condition across all four settings, $\SEL$ decreases from $.204$ at eight tokens to $.102$ at 64. The paired difference across source images is $.102$, with a 95\% confidence interval of [$.057$, $.150$]. Detailed settings and results are provided in the supplementary.

These attenuation trends are qualitatively consistent with prior analyses in which sensitivity to individual tokens or attention concentration on relevant information can weaken as input size grows \cite{barbero2024glasses,gollapudi2026retrieval,hahn2020theoretical,velickovic2024softmax}. This connection, however, does not establish $\SEL$'s functional form or identify the attenuation mechanism.

\subsection{Edit responsiveness is observed without precise masks}
\label{sec:load}

\textbf{Object tokens with positional information remain responsive to edits even without pooling within precise object boundaries.} We examine the role of boundary precision through two comparisons. First, we dilate masks by up to two feature-map cells and retrain the projector at each step. Of the regions added by dilation, approximately 17\% overlap other object masks on average, exceeding 30\% in the highest tenth. Oracle deletion $\SEL$ shows no clear decrease with dilation, with a slope of $.004$ [$-.008$, $.017$] across steps. Neither the residual presence probability after deletion nor token dependence in standard VQA changes consistently.

Pooling features from detection boxes instead of segmentation masks likewise yields no clear differences from Det+SAM in the evaluated metrics. The paired box-minus-mask difference in deletion $\SEL$ is $-.012$ on iSAID and $-.003$ on VRSBench. On VRSBench, boxes have a mean area 1.92 times that of their corresponding masks, so the difference in included surroundings is substantial. These results show edit responsiveness without precise boundaries under the tested conditions, complementing the analysis of how objects are separated into tokens in \cref{sec:cleanliness-density}. With a fixed detection vocabulary, box pooling may reduce computation; masks may serve a separate role for regions outside that vocabulary or specified by users. We nonetheless retain mask pooling as the default, providing a consistent reference for characterizing object-level representations.

\subsection{VQA performance is largely preserved, and responses depend on the tokens}
\label{sec:reading}

\begin{table}[t]
  \centering\footnotesize
  \caption{\textbf{Free-form VQA: preservation and token dependence.} LLM-judge scored; preservation is relative to the matched \armPatch{} reference. Normal and shuffled accuracies pool items, whereas driven averages within-run accuracy differences.}
  \label{tab:vqa}
  \setlength{\tabcolsep}{3pt}
  \begin{tabular}{llcccc}
    \toprule
    Arm & Train & Norm. & Shuf. & driven {\scriptsize[CI]} & Pres. \\
    \midrule
    \multicolumn{6}{l}{\scriptsize iSAID (judge: Qwen2.5-72B; 17{,}812 verdicts; \armPatch{} $.564$)} \\
    \armOracle & \ea & .541 & .382 & $+.159$ {\scriptsize$[.137,.186]$} & 96\% \\
    \armSeg & \ea & .527 & .381 & $+.145$ {\scriptsize$[.107,.178]$} & 93\% \\
    \armRaw & \ea & .526 & .380 & $+.144$ {\scriptsize$[.110,.175]$} & 93\% \\
    \armOracle & \vo & .517 & .361 & $+.156$ {\scriptsize$[.117,.192]$} & 92\% \\
    \armSeg & \vo & .527 & .353 & $+.175$ {\scriptsize$[.147,.202]$} & 94\% \\
    \armRaw & \vo & .529 & .385 & $+.146$ {\scriptsize$[.109,.178]$} & 94\% \\
    \armPatch & \vo & .564 & .377 & $+.187$ {\scriptsize$[.159,.214]$} & 100\% \\
    \midrule
    \multicolumn{6}{l}{\scriptsize VRSBench (16{,}640 verdicts; \armPatch{} $.528$)} \\
    \armSeg & \ea & .533 & .341 & $+.191$ {\scriptsize$[.161,.218]$} & 101\% \\
    \armRaw & \ea & .487 & .379 & $+.107$ {\scriptsize$[.088,.123]$} & 92\% \\
    \armSeg & \vo & .469 & .361 & $+.107$ {\scriptsize$[.088,.127]$} & 89\% \\
    \armRaw & \vo & .445 & .397 & $+.049$ {\scriptsize$[.024,.077]$} & 84\% \\
    \armPatch & \vo & .528 & .444 & $+.084$ {\scriptsize$[.044,.116]$} & 100\% \\
    \bottomrule
  \end{tabular}
\end{table}

\textbf{The object-token route retains a substantial fraction of standard VQA performance relative to the matched Patch reference.} Scoring free-form responses, it preserves 92--96\% of Patch accuracy ($.564$) on iSAID and 84--101\% of Patch accuracy ($.528$) on VRSBench. Det+SAM EA reaches 101\% on VRSBench (\cref{tab:vqa}). Shuffling the pairing between images and questions reduces performance: $\mathrm{driven}$ ranges from $.144$ to $.175$ across iSAID configurations, with all intervals above zero. All configurations on VRSBench also have $\mathrm{driven}$ intervals above zero. Responses therefore depend on correctly pairing object tokens with questions. On iSAID, VO shows this input dependence despite near-zero deletion $\SEL$, motivating separate evaluation of standard VQA and edit responsiveness. The relationship between the two measures varies by dataset: $\mathrm{driven}$ occupies a similar range across training conditions and token constructions on iSAID, but varies with both on VRSBench.

Providing the unedited image through the model's native visual route alongside object tokens reduces edit selectivity by 76--81\% and brings $\mathrm{driven}$ close to zero with the projector held fixed. Training with both images and tokens provides only limited recovery (supplementary). This shows that, when the original image is also present, object-token edits and correct token--question pairing can have less influence on the response.

\subsection{Effects of training and token construction recur across backbones}
\label{sec:universality}

\begin{table}[t]
  \centering
  \caption{\textbf{Training and token-construction contrasts across language backbones.} Values are $\SEL$ contrasts except the last row; $\star$ marks a 95\% CI excluding zero. The separate 0.5B comparison is reported in the text.}
  \label{tab:universality}
  \setlength{\tabcolsep}{2pt}
  \scriptsize
  \begin{tabular}{@{}lccc@{}}
    \toprule
    Comparison & LLaVA-OV-7B & Qwen2.5-VL & Idefics3-8B \\
    \midrule
    \multicolumn{4}{@{}l}{\scriptsize iSAID} \\
    \armOracle$-$\armRaw{} (\ea) & $+.050^\star$ & $+.066^\star$ & $+.060^\star$ \\
    Training \armOracle{} \ea$-$\vo & $+.047^\star$ & $+.063^\star$ & $+.050^\star$ \\
    Training \armSeg{} \ea$-$\vo & $+.041^\star$ & $+.033^\star$ & $+.031^\star$ \\
    \midrule
    \multicolumn{4}{@{}l}{\scriptsize VRSBench} \\
    Training \armSeg{} \ea$-$\vo & $+.425^\star$ & $+.448^\star$ & $+.406^\star$ \\
    \armSeg$-$\armRaw{} (\ea) & $+.446^\star$ & $+.444^\star$ & $+.320^\star$ \\
    Free-text driven $>0$ & \checkmark{} (\cref{tab:vqa}) & $+.176^\star$ & $+.142^\star$ \\
    \bottomrule
  \end{tabular}
\end{table}

\textbf{The main comparisons retain their direction across backbones with different training histories and language-model lineages.} We examine this by replacing the language backbone with Qwen2.5-VL-7B-Instruct \cite{bai2025qwen25vl} (a different backbone training history) and Idefics3-8B \cite{laurencon2024idefics3} (a different LM lineage), and retraining the projector. Under the same measurement procedure, all three 7B--8B backbones show an increase in deletion responsiveness with edit training and differences between token constructions in the same direction, with the corresponding contrast intervals above zero (\cref{tab:universality}). The pattern recurs on VRSBench, where VQA preservation relative to each backbone's Patch reference is 101\% for LLaVA-OV, 111\% for Qwen2.5-VL, and 93\% for Idefics3. All three backbones also have $\mathrm{driven}$ intervals above zero for free-form VQA. Effect sizes differ, but the directions of the repeated training and token-construction comparisons are preserved.

At 0.5B within the LLaVA-OV family, the additional responsiveness induced by edit training is very limited. Differences between token constructions retain their direction. Oracle shows no clear training effect (EA--VO CI includes zero), whereas Det+SAM retains a small positive effect. Note, however, that the projector also becomes smaller along with the language backbone in this comparison.

\subsection{Main trends persist across scoring choices}
\label{sec:designeffect}

\begin{table}[t]
  \centering\footnotesize
  \caption{\textbf{Sensitivity to scoring choices.} The first numeric column is a between-condition $\SEL$ contrast under the default scoring. Other columns are paired changes in that contrast after reversing one choice, over the same nine runs; $\star$ marks a 95\% CI excluding zero. Pre-edit correctness is changed by rescoring the same models.}
  \label{tab:design}
  \setlength{\tabcolsep}{2pt}
  \scriptsize
  \begin{tabular}{@{}lccc@{}}
    \toprule
    Comparison & $\SEL$ contrast & \begin{tabular}{@{}c@{}}$\Delta$ no\\placebo\end{tabular} & \begin{tabular}{@{}c@{}}$\Delta$ no pre-\\correct\end{tabular} \\
    \midrule
    EA--VO \armOracle & $+.047$ {\scriptsize$[.040,.055]$} & $+.002$ & $-.013^\star$ \\
    EA--VO \armSeg & $+.041$ {\scriptsize$[.031,.051]$} & $+.008$ & $-.016^\star$ \\
    \armOracle$-$\armRaw & $+.050$ {\scriptsize$[.043,.057]$} & $+.009^\star$ & $-.013^\star$ \\
    \armSeg$-$\armRaw & $+.044$ {\scriptsize$[.034,.054]$} & $+.014^\star$ & $-.016^\star$ \\
    \armOracle$-$\armSeg{} gap & $+.005$ {\scriptsize$[-.007,.018]$} & $-.005$ & $+.004$ \\
    \bottomrule
  \end{tabular}
\end{table}

\textbf{Across the scoring settings examined, magnitudes change but the main training and token-construction comparisons retain their direction (\cref{tab:design}).} We vary scoring choices for deletion, which has the largest evaluation set. Omitting the subtraction of control-edit effects increases the Oracle--SAMraw and Det+SAM--SAMraw $\SEL$ contrasts by $.009$ and $.014$, respectively. In contrast, the changes in the EA--VO contrasts are $.002$ and $.008$, with intervals that include zero. This indicates that the effects of deleting question-irrelevant tokens can differ across token constructions and should be measured separately for each condition. Without pre-edit correctness filtering, these contrasts decrease by $.013$--$.016$.

Under EA, Oracle and Det+SAM answer 41--44\% of candidate deletion questions incorrectly before editing, compared with 9--17\% under VO. A tendency to answer ``yes'' indiscriminately can also pass the pre-edit correctness condition on questions about objects that are present, so eligibility and edit responsiveness should be distinguished. Restricting the comparison to cases answered correctly before editing by both EA and VO preserves the training contrast: $.040$ [$.030$, $.051$] for Oracle and $.043$ [$.028$, $.061$] for Det+SAM. Scoring answer-choice flips instead of probability changes also preserves the direction of the main contrasts, with between-condition differences of 9.7--13.8 percentage points. The Oracle--Det+SAM interval includes zero under every scoring setting examined. The training effect also persists without an explicit yes/no instruction: deletion and addition answer-choice flip rates are 50--75\% with edit training and 0--2\% under VO (supplementary). As noted in \cref{sec:setup}, however, these values were obtained without per-condition tuning; we focus on comparative trends rather than absolute magnitudes.

% Section 5: approved English v3, 2026-10-26.
% Based on Korean v1 and English v2; plausible explanation / possible interpretation.
% The alternative comparison is provided in supplementary Sec. S4.13 (2026-11-02).
\section{Discussion and Limitations}
\label{sec:discussion}

\paragraph{The boundary map.} The results in \cref{sec:experiments} outline the conditions under which edit responsiveness emerges and the scope of these findings. \textbf{Query-type boundary:} For counting questions, $\mathrm{driven}$ pooled across object-token and image-patch settings is $+.016\,[-.000,+.032]$ (51 experimental conditions). Counting responses show no clear dependence on visual input in the tested configurations, suggesting that limitations in reading visual information can arise before edit responsiveness is considered. \textbf{Teaching boundary:} Addition responsiveness tends to be more restricted to classes included in edit training than deletion or movement responsiveness. Deletion and movement responses weaken but generally persist on excluded classes. A plausible explanation is that deletion and movement read class information from existing tokens built by the perception pipeline, an ability that may partially transfer to excluded classes. Addition instead requires treating an inserted token as evidence of class presence and may depend more on class-specific edit training. \textbf{Domain boundary:} Our observation that edit responsiveness can occur without precise object boundaries applies to the datasets and pooling conditions examined (\cref{sec:load}). This has not yet been verified in natural scenes with greater mutual occlusion, where precise segmentation masks may play a larger role. \textbf{Capacity boundary:} In the 0.5B model from the LLaVA-OV family, the additional responsiveness induced by edit training is very limited (\cref{sec:universality}). One possible interpretation is that the frozen language model already has abilities for reasoning about position and existence, and edit training aligns the object representation with those abilities.

\paragraph{Limitations of the measurement.} Our measurement has several limitations. First, we simulate user operations programmatically, without validation with real users or in practical settings. Second, the main quantitative analyses focus on targeted queries with predefined answer candidates (\cref{sec:protocol}). We also examine how edits are reflected in free-form responses in the supplementary, but broader evaluation without uniquely determined answers requires further study. Third, determining that the target class is absent from a region for addition relies on ground-truth annotations. The pre-addition judgment of absence assumes that all objects of that class in the region have been annotated. Fourth, our findings are based on the object-token testbed and the training and input conditions examined here. Whether the same characteristics hold for other representations requires further validation.

\paragraph{Comparison with alternatives and future work.} We also report a small-scale comparison with textual hypothetical prompting and generative image editing in supplementary Sec.~S4.13. No consistent performance advantage for the object-token approach was observed under the evaluated conditions. However, the comparison should be interpreted with differences in the approaches' operation and characteristics in mind. Our object-token models use the existing analysis checkpoints and training settings, without additional tuning for the comparison tasks. Exploring settings to improve performance and conducting broader comparisons remain future work.

% Research question aligned with the confirmed title, 2026-10-26.
% Section 6: approved English v3, 2026-10-26.
% iSAID follows the matched patch-token baseline; question/When answer retained.
\section{Conclusion}
\label{sec:conclusion}

When do frozen VLMs respond to image-free object-token edits? \textbf{When trained with edit examples, they respond more strongly to deletion, movement, and addition, with responsiveness shaped by object-token construction, context token count, and model capacity.} The directions of the main training and token-construction contrasts recur across two remote-sensing datasets and three frozen language backbones. Standard VQA accuracy is preserved at $92$--$96\%$ of a matched patch-token baseline on iSAID, and responses remain dependent on the object tokens. The training and token-construction contrasts examined in our robustness checks retain their direction across scoring and evaluation settings. We release code, evaluation-question generation rules, measurement records, and scoring logs to support reproducible analyses and follow-up research. Through systematic analysis, we show when frozen VLMs can respond to hypothetical changes applied directly to object-token inputs, and we hope these findings will guide object-token editing systems and related research.

\label{cvpr:body-end}

% LLM usage disclosure revised and approved by the user, 2026-11-06.
\section*{LLM Usage Statement}
\label{sec:llm-usage}
\noindent We used OpenAI Codex and Claude Code to assist with manuscript drafting and editing, literature search and reference checking, organization of the supplementary material, coding, and analysis. We used Qwen2.5-72B-Instruct-AWQ and GPT-4o for automated response evaluation, as described in Supplementary Secs. S2.6 and S5, and Qwen-Image-Edit-2511 for the generative image-editing comparison in Supplementary Sec. S4.13. The authors take full responsibility for the final content.

{\small
  \bibliographystyle{ieeenat_fullname}
  \bibliography{main}

@String(CVPR= {IEEE Conf. Comput. Vis. Pattern Recog.})

@String(CVPRW= {IEEE Conf. Comput. Vis. Pattern Recog. Worksh.})

@String(ICCV= {Int. Conf. Comput. Vis.})

@String(ECCV= {Eur. Conf. Comput. Vis.})

@String(NIPS= {Adv. Neural Inform. Process. Syst.})

@String(ICLR= {Int. Conf. Learn. Represent.})

@String(CORL= {Conf. Robot Learn.})

@String(TACL= {Trans. Assoc. Comput. Linguistics})

@String(ACMMM= {ACM Int. Conf. Multimedia})

@String(ICML= {Int. Conf. Mach. Learn.})

@String(TMLR= {Trans. Mach. Learn. Res.})

@String(TGRS= {IEEE Trans. Geosci. Remote Sens.})

@inproceedings{zamir2019isaid,
  author    = {Waqas Zamir, Syed and Arora, Aditya and Gupta, Akshita and Khan, Salman and Sun, Guolei and Shahbaz Khan, Fahad and Zhu, Fan and Shao, Ling and Xia, Gui-Song and Bai, Xiang},
  title     = {{iSAID}: A Large-scale Dataset for Instance Segmentation in Aerial Images},
  booktitle = CVPRW,
  pages     = {28--37},
  year      = {2019},
}

@inproceedings{xia2018dota,
  author    = {Xia, Gui-Song and Bai, Xiang and Ding, Jian and Zhu, Zhen and Belongie, Serge and Luo, Jiebo and Datcu, Mihai and Pelillo, Marcello and Zhang, Liangpei},
  title     = {{DOTA}: A Large-scale Dataset for Object Detection in Aerial Images},
  booktitle = CVPR,
  year      = {2018},
}

@inproceedings{li2024vrsbench,
  author    = {Li, Xiang and Ding, Jian and Elhoseiny, Mohamed},
  title     = {{VRSBench}: A Versatile Vision-Language Benchmark Dataset for Remote Sensing Image Understanding},
  booktitle = NIPS,
  year      = {2024},
}

@inproceedings{zi2025rsvlmqa,
  author    = {Zi, Xing and Xiao, Jinghao and Shi, Yunxiao and Tao, Xian and Li, Jun and Braytee, Ali and Prasad, Mukesh},
  title     = {{RSVLM-QA}: A Benchmark Dataset for Remote Sensing Vision Language Model-based Question Answering},
  booktitle = ACMMM,
  pages     = {12905--12911},
  year      = {2025},
}

@article{li2024llavaonevision,
  author    = {Li, Bo and Zhang, Yuanhan and Guo, Dong and Zhang, Renrui and Li, Feng and Zhang, Hao and Zhang, Kaichen and Zhang, Peiyuan and Li, Yanwei and Liu, Ziwei and Li, Chunyuan},
  title     = {{LLaVA-OneVision}: Easy Visual Task Transfer},
  journal   = TMLR,
  year      = {2025},
}

@misc{yu2026mindedit,
  author    = {Yu, Leyuan and Tang, Xiao and Liu, Minghao and Li, Xinyuan and Bai, Xiaokai and Zhou, Sheng and Lin, Qunshu and Xuan, Weihao and Yokoya, Naoto},
  title     = {{MindEdit-Bench}: Benchmarking Object-Level Counterfactual Spatial Reasoning in {VLMs} from In-the-Wild Photos},
  eprint    = {2607.00491},
  archivePrefix = {arXiv},
  note      = {arXiv:2607.00491},
  year      = {2026},
}

@misc{bai2025qwen25vl,
  author    = {Bai, Shuai and Chen, Keqin and Liu, Xuejing and Wang, Jialin and Ge, Wenbin and Song, Sibo and Dang, Kai and Wang, Peng and Wang, Shijie and Tang, Jun and Zhong, Humen and Zhu, Yuanzhi and Yang, Mingkun and Li, Zhaohai and Wan, Jianqiang and Wang, Pengfei and Ding, Wei and Fu, Zheren and Xu, Yiheng and Ye, Jiabo and Zhang, Xi and Xie, Tianbao and Cheng, Zesen and Zhang, Hang and Yang, Zhibo and Xu, Haiyang and Lin, Junyang},
  title     = {{Qwen2.5-VL} Technical Report},
  eprint    = {2502.13923},
  archivePrefix = {arXiv},
  note      = {arXiv:2502.13923},
  year      = {2025},
}

@misc{laurencon2024idefics3,
  author    = {Lauren{\c{c}}on, Hugo and Marafioti, Andr{\'e}s and Sanh, Victor and Tronchon, L{\'e}o},
  title     = {Building and Better Understanding Vision-Language Models: Insights and Future Directions},
  eprint    = {2408.12637},
  archivePrefix = {arXiv},
  note      = {arXiv:2408.12637},
  year      = {2024},
}

@misc{qwen2024qwen25,
  author    = {{Qwen Team}},
  title     = {{Qwen2.5} Technical Report},
  eprint    = {2412.15115},
  archivePrefix = {arXiv},
  note      = {arXiv:2412.15115},
  year      = {2024},
}

@inproceedings{zhai2023siglip,
  author    = {Zhai, Xiaohua and Mustafa, Basil and Kolesnikov, Alexander and Beyer, Lucas},
  title     = {Sigmoid Loss for Language Image Pre-Training},
  booktitle = ICCV,
  year      = {2023},
}

@inproceedings{liu2022convnext,
  author    = {Liu, Zhuang and Mao, Hanzi and Wu, Chao-Yuan and Feichtenhofer, Christoph and Darrell, Trevor and Xie, Saining},
  title     = {A {ConvNet} for the 2020s},
  booktitle = CVPR,
  year      = {2022},
}

@misc{ilharco2021openclip,
  author    = {Ilharco, Gabriel and Wortsman, Mitchell and Wightman, Ross and Gordon, Cade and Carlini, Nicholas and Taori, Rohan and Dave, Achal and Shankar, Vaishaal and Namkoong, Hongseok and Miller, John and Hajishirzi, Hannaneh and Farhadi, Ali and Schmidt, Ludwig},
  title     = {{OpenCLIP}},
  year      = {2021},
  note      = {Zenodo. doi:10.5281/zenodo.5143773},
}

@inproceedings{kirillov2023sam,
  author    = {Kirillov, Alexander and Mintun, Eric and Ravi, Nikhila and Mao, Hanzi and Rolland, Chloe and Gustafson, Laura and Xiao, Tete and Whitehead, Spencer and Berg, Alexander C. and Lo, Wan-Yen and Doll{\'a}r, Piotr and Girshick, Ross},
  title     = {Segment Anything},
  booktitle = ICCV,
  year      = {2023},
}

@misc{jocher2024yolo11,
  author    = {Jocher, Glenn and Qiu, Jing},
  title     = {Ultralytics {YOLO11}},
  year      = {2024},
  note      = {\url{https://github.com/ultralytics/ultralytics}},
}

@inproceedings{liu2025rsthinker,
  author    = {Liu, Jiaqi and Sun, Lang and Fu, Ronghao and Yang, Bo},
  title     = {Towards Faithful Reasoning in Remote Sensing: A Perceptually-Grounded {GeoSpatial} Chain-of-Thought for Vision-Language Models},
  booktitle = ICLR,
  year      = {2026},
}

@misc{luo2024skysensegpt,
  author    = {Luo, Junwei and Pang, Zhen and Zhang, Yongjun and Wang, Tingzhu and Wang, Linlin and Dang, Bo and Lao, Jiangwei and Wang, Jian and Chen, Jingdong and Tan, Yihua and Li, Yansheng},
  title     = {{SkySenseGPT}: A Fine-Grained Instruction Tuning Dataset and Model for Remote Sensing Vision-Language Understanding},
  eprint    = {2406.10100},
  archivePrefix = {arXiv},
  note      = {arXiv:2406.10100},
  year      = {2024},
}

@inproceedings{kuckreja2024geochat,
  author    = {Kuckreja, Kartik and Danish, Muhammad Sohail and Naseer, Muzammal and Das, Abhijit and Khan, Salman and Khan, Fahad Shahbaz},
  title     = {{GeoChat}: Grounded Large Vision-Language Model for Remote Sensing},
  booktitle = CVPR,
  year      = {2024},
}

@article{elgendy2025chatenv,
  author    = {Elgendy, Hosam and Sharshar, Ahmed and Aboeitta, Ahmed and Guizani, Mohsen},
  title     = {{ChatENV}: An Interactive Vision-Language Model for Sensor-Guided Environmental Monitoring and Scenario Simulation},
  journal   = TGRS,
  volume    = {64},
  pages     = {4703710},
  year      = {2026},
}

@inproceedings{chappuis2022promptrsvqa,
  author    = {Chappuis, Christel and Zermatten, Val{\'e}rie and Lobry, Sylvain and Le Saux, Bertrand and Tuia, Devis},
  title     = {Prompt-{RSVQA}: Prompting Visual Context to a Language Model for Remote Sensing Visual Question Answering},
  booktitle = CVPRW,
  year      = {2022},
}

@misc{jahagirdar2026maskllava,
  author    = {Jahagirdar, Soumya and Bousselham, Walid and Kukleva, Anna and Kuehne, Hilde},
  title     = {When {LLaVA} Meets Objects: Token Composition for Vision-Language-Models},
  eprint    = {2602.04864},
  archivePrefix = {arXiv},
  note      = {arXiv:2602.04864},
  year      = {2026},
}

@misc{zhang2025adatok,
  author    = {Zhang, Xinliang and Zhu, Lei and He, Hangzhou and Zeng, Shuang and Fu, Ourui and Hu, Jiakui and Yao, Zhengjian and Lu, Yanye},
  title     = {{AdaTok}: Adaptive Token Compression with Object-Aware Representations for Efficient Multimodal {LLMs}},
  eprint    = {2511.14169},
  archivePrefix = {arXiv},
  note      = {arXiv:2511.14169},
  year      = {2025},
}

@inproceedings{tian2024tokenav,
  author    = {Tian, Ran and Li, Boyi and Weng, Xinshuo and Chen, Yuxiao and Schmerling, Edward and Wang, Yue and Ivanovic, Boris and Pavone, Marco},
  title     = {Tokenize the World into Object-level Knowledge to Address Long-tail Events in Autonomous Driving},
  booktitle = CORL,
  year      = {2024},
}

@inproceedings{huang2024chatscene,
  author    = {Huang, Haifeng and Chen, Yilun and Wang, Zehan and Huang, Rongjie and Xu, Runsen and Wang, Tai and Liu, Luping and Cheng, Xize and Zhao, Yang and Pang, Jiangmiao and Zhao, Zhou},
  title     = {Chat-Scene: Bridging {3D} Scene and Large Language Models with Object Identifiers},
  booktitle = NIPS,
  year      = {2024},
}

@inproceedings{yuan2024osprey,
  author    = {Yuan, Yuqian and Li, Wentong and Liu, Jian and Tang, Dongqi and Luo, Xinjie and Qin, Chi and Zhang, Lei and Zhu, Jianke},
  title     = {Osprey: Pixel Understanding with Visual Instruction Tuning},
  booktitle = CVPR,
  year      = {2024},
}

@inproceedings{wu2024setok,
  author    = {Wu, Shengqiong and Fei, Hao and Li, Xiangtai and Ji, Jiayi and Zhang, Hanwang and Chua, Tat-Seng and Yan, Shuicheng},
  title     = {Towards Semantic Equivalence of Tokenization in Multimodal {LLM}},
  booktitle = ICLR,
  year      = {2025},
}

@inproceedings{zhang2023cvqa,
  author    = {Zhang, Letian and Zhai, Xiaotong and Zhao, Zhongkai and Zong, Yongshuo and Wen, Xin and Zhao, Bingchen},
  title     = {What If the {TV} Was Off? Examining Counterfactual Reasoning Abilities of Multi-modal Language Models},
  booktitle = CVPR,
  year      = {2024},
}

@inproceedings{agarwal2020ivvqa,
  author    = {Agarwal, Vedika and Shetty, Rakshith and Fritz, Mario},
  title     = {Towards Causal {VQA}: Revealing and Reducing Spurious Correlations by Invariant and Covariant Semantic Editing},
  booktitle = CVPR,
  year      = {2020},
}

@inproceedings{gupta2022swapmix,
  author    = {Gupta, Vipul and Li, Zhuowan and Kortylewski, Adam and Zhang, Chenyu and Li, Yingwei and Yuille, Alan},
  title     = {{SwapMix}: Diagnosing and Regularizing the Over-Reliance on Visual Context in Visual Question Answering},
  booktitle = CVPR,
  year      = {2022},
}

@inproceedings{yebin2024beaf,
  author    = {Ye-Bin, Moon and Hyeon-Woo, Nam and Choi, Wonseok and Oh, Tae-Hyun},
  title     = {{BEAF}: Observing {BE}fore-{AF}ter Changes to Evaluate Hallucination in Vision-Language Models},
  booktitle = ECCV,
  year      = {2024},
}

@misc{ding2025edct,
  author    = {Ding, Sihao and Vasa, Santosh and Ramadwar, Aditi},
  title     = {Explanation-Driven Counterfactual Testing for Faithfulness in Vision-Language Model Explanations},
  eprint    = {2510.00047},
  archivePrefix = {arXiv},
  year      = {2025},
  note      = {NeurIPS Workshop on Regulatable ML. arXiv:2510.00047},
}

@inproceedings{jiang2025interpreting,
  author    = {Jiang, Nicholas and Kachinthaya, Anish and Petryk, Suzanne and Gandelsman, Yossi},
  title     = {Interpreting and Editing Vision-Language Representations to Mitigate Hallucinations},
  booktitle = ICLR,
  year      = {2025},
}

@inproceedings{niu2021cfvqa,
  author    = {Niu, Yulei and Tang, Kaihua and Zhang, Hanwang and Lu, Zhiwu and Hua, Xian-Sheng and Wen, Ji-Rong},
  title     = {Counterfactual {VQA}: A Cause-Effect Look at Language Bias},
  booktitle = CVPR,
  year      = {2021},
}

@inproceedings{zarlenga2023intcem,
  author    = {Espinosa Zarlenga, Mateo and Collins, Katherine M. and Dvijotham, Krishnamurthy and Weller, Adrian and Shams, Zohreh and Jamnik, Mateja},
  title     = {Learning to Receive Help: Intervention-Aware Concept Embedding Models},
  booktitle = NIPS,
  year      = {2023},
}

@inproceedings{laguna2024intervenable,
  author    = {Laguna, Sonia and Marcinkevi{\v{c}}s, Ri{\v{c}}ards and Vandenhirtz, Moritz and Vogt, Julia E.},
  title     = {Beyond Concept Bottleneck Models: How to Make Black Boxes Intervenable?},
  booktitle = NIPS,
  year      = {2024},
}

@inproceedings{li2025maskadapter,
  author    = {Li, Yongkang and Cheng, Tianheng and Feng, Bin and Liu, Wenyu and Wang, Xinggang},
  title     = {Mask-Adapter: The Devil is in the Masks for Open-Vocabulary Segmentation},
  booktitle = CVPR,
  year      = {2025},
}

@article{hahn2020theoretical,
  author    = {Hahn, Michael},
  title     = {Theoretical Limitations of Self-Attention in Neural Sequence Models},
  journal   = TACL,
  volume    = {8},
  pages     = {156--171},
  year      = {2020},
}

@inproceedings{barbero2024glasses,
  author    = {Barbero, Federico and Banino, Andrea and Kapturowski, Steven and Kumaran, Dharshan and Ara{\'u}jo, Jo{\~a}o G.M. and Vitvitskyi, Alex and Pascanu, Razvan and Veli{\v{c}}kovi{\'c}, Petar},
  title     = {Transformers Need Glasses! Information Over-squashing in Language Tasks},
  booktitle = NIPS,
  year      = {2024},
}

@inproceedings{velickovic2024softmax,
  author    = {Veli{\v{c}}kovi{\'c}, Petar and Perivolaropoulos, Christos and Barbero, Federico and Pascanu, Razvan},
  title     = {Softmax Is Not Enough (for Sharp Size Generalisation)},
  booktitle = ICML,
  year      = {2025},
}

@misc{gollapudi2026retrieval,
  author    = {Gollapudi, Siddharth and Gupta, Nilesh and Singhal, Prasann and Min, Sewon},
  title     = {Can Language Models Actually Retrieve In-Context? Drowning in Documents at Million Token Scale},
  eprint    = {2607.01538},
  archivePrefix = {arXiv},
  note      = {arXiv:2607.01538},
  year      = {2026},
}
}
\end{document}